\documentclass[sigconf]{acmart}
\AtBeginDocument{%
  }

\usepackage{listings}
\usepackage{booktabs}
\usepackage{tikz}
\usepackage{multirow}

\usetikzlibrary{positioning}

\setcopyright{acmlicensed}

\copyrightyear{2026}
\acmYear{2026}
\setcopyright{cc}
\setcctype{by}
\acmConference[CAIS '26]{ACM Conference on AI and Agentic Systems}{May 26--29, 2026}{San Jose, CA, USA}
\acmBooktitle{ACM Conference on AI and Agentic Systems (CAIS '26), May 26--29, 2026, San Jose, CA, USA}
\acmDOI{10.1145/3786335.3813192}
\acmISBN{979-8-4007-2415-2/2026/05}

\begin{document}

\title{Governance by Construction for Generalist Agents}




\settopmatter{authorsperrow=5}

\newcommand{\ibmaffiliation}{%
\affiliation{%
  \institution{IBM}
  \city{Haifa}
  \country{Israel}
}
}

\author{Segev Shlomov}\authornote{Corresponding author email: segev.shlomov1@ibm.com.}
\ibmaffiliation

\author{Iftach Shoham}
\ibmaffiliation

\author{Alon Oved}
\ibmaffiliation

\author{Ido Levy}
\ibmaffiliation

\author{Sami Marreed}
\ibmaffiliation

\author{Harold Ship}
\ibmaffiliation

\author{Offer Akrabi}
\ibmaffiliation

\author{Sergey Zeltyn}
\ibmaffiliation

\author{Avi Yaeli}
\ibmaffiliation

\author{Nir Mashkif}
\ibmaffiliation
\renewcommand{\shortauthors}{Shlomov et~al.}

\begin{abstract}
Enterprise agents are increasingly expected to operate autonomously across tools and interfaces, yet production deployments require governance by construction. Systems must specify which actions are allowed, when human oversight is required, and what information may be exposed, without rebuilding the agent for each domain. This demo presents CUGA's policy system, a modular policy-as-code layer that composes with a generalist LLM agent to deliver predictable, auditable, and compliance-aware behavior in compound workflows without model fine-tuning. We present a runtime governance architecture that enforces policy interventions at every critical stage of execution. Rather than passively constraining behavior, policies intercept the agent at five structural checkpoints: upstream of planning (Intent Guard), within the system prompt to steer reasoning (Playbook), at the tool-call boundary to enforce proper usage (Tool Guide), outside the reasoning loop as a Human-in-the-Loop gate for high-risk actions (Tool Approvals), and at the output stage to filter and structure the final response (Output Formatter). Together, these stages embed governance continuously across the agent's execution pipeline rather than treating it as an afterthought. Using a healthcare scenario and a multi-layered enforcement intervention, the demo shows dynamic playbook injection for structured tool-sequence enforcement, intent guards that block malicious or accidental harmful requests, and human-in-the-loop tool approval checkpoints for potentially destructive actions. The artifact illustrates how typed governance primitives enable faster, safer deployment of enterprise agentic systems while improving policy adherence and execution consistency.
\end{abstract}



\maketitle

\section{Introduction}
\label{sec:intro}

LLM–based agents are increasingly adapted to perform complex, multi-step tasks across enterprise software environments, extending earlier work on conversational RPA, next-action recommendation, and process-aware automation~\cite{yaeli2022recommending,zeltyn2022prescriptive,oved2025snap}. Unlike static chatbots, modern agents integrate planning, internal and external tool-use, memory, and iterative reasoning to execute compound workflows across heterogeneous systems. Recent advances in tool-augmented and computer-using agents enable interaction with APIs, databases, and user interfaces, allowing agents to autonomously retrieve data, modify records and trigger communications. Although this flexibility enables generalization, prior work on LLM automation agents and web-agent benchmarks~\cite{ying2025securewebarena} suggests~\cite {ma2026safety,yang2025mla,luo2025agentauditor,shi2025towards,jiang2025think,chen2025shieldagent} that it also introduces unpredictability: agents may hallucinate facts, misuse tools, violate procedural constraints, behave inconsistently, or expose sensitive information~\cite{schwartz2023enhancing,shlomov2025grounding,levy2024st}. In enterprise settings, such failures are not merely quality degradations; they can result in compliance violations, data leakage, financial impacts, or reputational damage.
 
Many current governance strategies rely on prompt-engineering techniques, instruction stuffing, constraint injection, and post-hoc validation~\cite{tsai2025contextual, gaurav2025governance, zwerdling2025towards}. While these approaches can shape model behavior in controlled scenarios, they exhibit several structural limitations. Behavioral constraints are tightly coupled to prompt structure and can become brittle as policies evolve; governance logic often becomes duplicated across multi-agent deployments; enforcement decisions are delegated to model reasoning and therefore difficult to audit; and independently implemented guardrails may produce inconsistent outcomes without principled conflict resolution. More structured approaches encode operating procedures through role specialization or guard code~\cite{hong2023metagpt, zwerdling2025towards}, but typically require substantial architectural commitment or domain-specific reconfiguration when procedures change. Parlant's~\cite{parlant2025} guideline‑driven alignment layers dynamically retrieve and inject natural-language rules and tool use at inference time but remain focused on local, chat-level compliance. Skill-based frameworks such as Claude's skill system provide reusable behavioral templates applied during execution. Both approaches still rely on the model to interpret and follow instructions, leaving policy adherence inherently probabilistic and vulnerable. As enterprise adoption accelerates, governance must evolve beyond prompt-level heuristics toward policy-by-construction: explicit, typed, runtime-enforced control primitives that operate independently of the model itself.

In this work, we present the implementation of our vision for trustworthy, enterprise-ready generalist agents, building on prior work on automation trust, web-agent bottlenecks, and safety-oriented agent evaluation, and demonstrate it through the open-source CUGA agent~\cite{shlomov2025benchmarks,marreed2025towards,cuga2026github}. We introduce the \textit{CUGA policy system}, a modular policy-as-code layer built into CUGA without requiring model fine-tuning. The framework introduces typed governance primitives that constrain intent recognition, planning, tool invocation, human approval requirements, and output formatting \emph{at runtime}, enforcing consistency and governance. Policies are matched using lightweight triggers (keywords and embedding similarity), which mitigates reliance on probabilistic methods, resolved through explicit conflict handling, and produces structured explanation traces for observability and quality assurance. It offers no need for architectural changes, complies using open-source models (e.g., \texttt{GPT-OSS-120B}), for on-prem deployment, does not affect or modify the agent's output, and showcases build-in observability. By externalizing governance into composable runtime policies, the system enables predictable, auditable, and compliance-aware behavior across compound workflows while preserving the flexibility of general-purpose LLM agents ensuring compliance and trustworthiness in enterprise-ready agents. We demonstrate our approach through an end-to-end enterprise workflow that highlights runtime policy enforcement. The demonstration centers on two representative scenarios: a healthcare assistance workflow and a multi-layered enforcement intervention. 

The demo showcases four out of five modular governance primitives defined as markdown policies at by interference at difference execution checkpoints. The Playbook enforces structured multi-step execution by dynamically associating requests with predefined tool sequences and enterprise-specific constraints, as the Tool Guide enriches tool description. The Intent Guard prevents malicious or accidental harmful actions by intercepting and blocking restricted intents prior to tool execution. The Tool Approval mechanism introduces a \textit{human-in-the-loop} (HITL) checkpoint that pauses the execution graph and requires explicit confirmation before potentially destructive actions can proceed.

\section{Methods Overview}
\label{sec:method_overview}

Our CUGA’s policy system strengthens enterprise-ready governance through dynamic interception of different policies throughout the agent's execution graph via the trigger mechanism in five different points: First, \emph{Intent Guard} sit at the very beginning of the process to immediately block bad requests before the agent even takes action. Second, \emph{Playbook} are seamlessly injected into the system prompt, step-by-step to steer the agent's planning and influence its reasoning. Third, \emph{Tool guide} update and mutate tool descriptions right before execution to instruct the agent on proper tool usage. Fourth, \emph{Tool approval} act as a critical safeguard outside the reasoning loop to gate execution, pausing the graph to wait for human confirmation if a risky action is attempted. Finally, \emph{Output formatter} serves as the last intervention point to appropriately filter the final answer before it is returned. This section presents the architectural design principles, data models, and runtime mechanisms underlying each policy category. The emphasis is placed on modularity, extensibility, and seamless integration with the LangGraph execution framework.

\subsection{Policy System Architecture}
\label{subsec:architecture}

The policy system is implemented as a modular, layered framework consisting of four architectural tiers:

\begin{enumerate}
    \item \textbf{Policy Models Layer}: Strongly-typed data models defining policy schemas, triggers, and action semantics.
    \item \textbf{Storage Layer}: Persistent and semantic storage backed by a vector database for similarity-based retrieval.
    \item \textbf{Policy Agent Layer}: Runtime matching and conflict-resolution logic that evaluates policies against execution context.
    \item \textbf{Enactment Layer}: Execution primitives that apply policy decisions within the LangGraph workflow.
\end{enumerate}

\noindent This layered design enforces separation of concerns: policy representation is decoupled from storage, matching logic, and execution semantics. Policy evaluation occurs at four semantically meaningful checkpoints: (1) intent analysis, (2) tool preparation, (3) post-code generation, and (4) final response generation.

\paragraph{\textbf{Trigger System.}}
\label{subsubsec:triggers}

All policy types, except for Tool Approval, rely on a \textit{configurable trigger mechanism}. Triggers are defined as discriminated union types, enabling flexible matching strategies within a unified interface. Supported trigger mechanisms include:

\begin{itemize}
    \item \textbf{Natural Language}: Semantic similarity matching via embedding-based retrieval with configurable similarity thresholds.
    \item \textbf{Keyword}: Exact or fuzzy keyword matching with logical composition (AND/OR) and case sensitivity controls.
    \item \textbf{Application}: Contextual matching based on the active application domain.
    \item \textbf{State}: Evaluation against structured agent state using equality, containment, or regular expression operators.
    \item \textbf{Tool}: Detection of tool usage at specific execution stages (pre- or post-invocation).
\end{itemize}

\noindent Triggers may target different contextual fields, including inferred intent, intermediate sub-tasks, and final agent responses. This flexibility enables policies to intervene at different abstraction levels of the execution life-cycle. Policy persistence and semantic retrieval are implemented using Milvus as a vector database backend. Embeddings are generated using either API or a local-based encoder~\cite{wang2020minilm}.

\subsection{Intent Guard}
\label{subsec:intent-guard}

Intent Guard policies enforce hard constraints by intercepting user intent outside the agent reasoning loop, before the agent acts. Their primary purpose is early-stage blocking of unauthorized or restricted operations. Upon activation, an Intent Guard terminates execution immediately, this early termination mechanism prevents downstream reasoning or unwanted tool invocation. Intent Guard trigger evaluation follows a two-phase process: \textbf{(1) Deterministic:} keyword-based triggers are evaluated first. Intent Guards are prioritized over other policy types to guarantee that blocking constraints supersede advisory policies. \textbf{(2) conflict resolution:} for natural-language triggers, multiple policies may satisfy the similarity threshold. In such cases, an LLM-based structured reasoning step selects the most appropriate policy. The model outputs a selected index, confidence score, and justification.

\subsection{Playbook}
\label{subsec:playbook}

Playbook policies provide structured workflow guidance for complex tasks. Rather than blocking execution, they shape agent planning behavior by injecting step-by-step instructions. This is particularly valuable as it delivers precise, targeted instructions without inflating the prompt with excessive tokens. This causes the agent to follow instructions and consistently comply to the user's tasks. The Playbook consists of markdown-formatted guidance content, an optional ordered list of steps, optional expected outcomes per step, optional tool constraints per step, trigger definitions, and priority. This structured representation enables fine-grained orchestration and verification of multi-step workflows.


\subsection{Tool Approval}
\label{subsec:tool-approval}

Tool Approval policies enforce \textit{HITL} oversight by requiring explicit confirmation before executing sensitive tools. Unlike trigger-based policies, they are evaluated after code generation, allowing inspection of the actual tools the agent intends to invoke in runtime.

After code generation, the system scans the code for tool invocations. If a matching tool is detected, execution pauses and enters a waiting state. The agent resumes only after explicit approval (or automatic approval, if configured). If multiple policies match, the highest-priority policy is applied. This mechanism is particularly useful for sensitive operations such as data modifications (e.g., database writes and updates) and external API calls, where interactions with third-party services should be reviewed before execution.

\subsection{Tool Guide}
\label{subsec:tool-guide}

Tool Guide policies augment tool descriptions with contextual or compliance-related guidance. Multiple Tool Guide policies may apply simultaneously, as they are cumulative rather than mutually exclusive. At runtime, tool definitions are deep-copied and enriched with the configured guidance, which may be appended to the original description. The deep-copy mechanism ensures that modifications remain session-scoped and do not permanently alter the underlying tool metadata.

\subsection{Output Formatter}
\label{subsec:output-formatter}

Output Formatter policies transform the final response into a structured format. They support three modes: (1) a predefined template verbatim, (2) restructuring the response as formatted Markdown, or (3) extracting structured data according to a specified JSON schema. Trigger evaluation considers both the user input and the generated response, enabling context-aware formatting decisions without relying on the agent's capabilities.

\section{System Demonstration: CUGA's Policy System}
\label{sec:system_demo}

The demo video\footnote{\url{https://www.youtube.com/watch?v=Ie5ZODIW5gk}} presents the live CUGA interface, illustrating how governance policies are enforced through targeted interventions in the agent execution graph. Rather than embedding complex enterprise constraints directly in prompts, CUGA applies policies dynamically during agent execution, enabling safer and more reliable task handling (for more information see Section~\ref{sec:method_overview}). Highlighting the importance of governance by construction, the walkthrough presents two representative scenarios designed to showcase CUGA’s ability to guide agent behavior while maintaining strong safety guarantees: (1) a complex healthcare workflow (OAK) and (2) a multi-layered security intervention.

\paragraph{\textbf{Playbook}}
The first scenario demonstrates CUGA in the context of a healthcare assistance task. A user issues the request \textit{``find primary care doctors near me.''}, CUGA dynamically associates the request with a predefined \textit{playbook} that enforces a structured execution policy for mandatory tool sequence to grantee reliable compliance. The execution begins with context extraction, where CUGA retrieves the user’s active insurance coverage and extracts key attributes required for downstream operations, including the relevant contract UID and brand code. Guided by the injected policy, the agent then invokes the \textit{find care suggestions} tool to map the natural language phrase ``primary care'' to the corresponding internal system code (code 25). This mapping ensures alignment with the underlying service taxonomy.

Following this step, the system performs a paginated provider search guided by the \textbf{Tool Guide} critical pagination requirement, issuing multiple API calls that collectively return 14 candidate providers. During the retrieval process, CUGA automatically applies a network-status constraint to ensure that only \textit{in-network} providers are considered. The results are then aggregated and presented to the user as a cleanly formatted table listing nearby in-network primary care physicians. This scenario demonstrates how CUGA is enabled to operate within a controlled policy framework while enforcing structured workflows, tool usage constraints, and enterprise-specific rules.

\paragraph{\textbf{Intent Guards and Tool Approval}}

The second scenario focuses on CUGA’s preventive capabilities, which introduces safety mechanisms that operate independently of the agent’s reasoning process. First, a user attempts to execute a malicious (which can also be by accidental) command by requesting to \textit{``delete every contact in CRM.''} the \textit{Intent Guard} immediately intercepts the request and issues a block command that terminates the execution graph. As a result, the agent never receives the instruction, preventing the agent from reasoning about or executing the harmful request. The demonstration then simulates a scenario in which an attacker paraphrases the request in a way that some how bypasses the initial guard. In this case, the agent proceeds to generate code intended to drop a database. Before the action can be executed, CUGA’s \textit{Tool Approval} mechanism pauses the execution graph and introduces a HITL checkpoint. The system requires explicit human confirmation before allowing the operation to proceed, ensuring that potentially destructive actions cannot be executed without oversight or accountability by the operator. Together, these scenarios highlight CUGA’s ability to combine policy-driven execution with layered safety mechanisms, enabling reliable and secure deployment of language agents in enterprise environments.

\section{Results}
\label{sec:results}

We evaluate CUGA's \textit{policy system} on two enterprise-focused generalist agent benchmarks that reflect the structured, policy-governed workflows central to our demonstration. \textbf{OAK}~\cite{oakbench2026github} is a 27-task customer-care benchmark grounded in realistic insurance scenarios, assessing the agent's ability to process claims, retrieve coverage information, and answer health-insurance questions via API calls. \textbf{BPO}~\cite{bpobench2026huggingface} is a 26-task business-process benchmark covering a broader range of enterprise back-office operations, providing a more challenging generalist evaluation. Together, these benchmarks reflect real-world enterprise settings where tasks carry well-defined procedures, tool-heavy execution, and low tolerance for deviation. All results are reported across three backbone models: \textit{GPT-OSS-120B}, our main focus as an open-source model, \textit{GPT-4.1}, and \textit{Claude Opus-4.5}. The primary metric is \textbf{Success Rate} (SR): the percentage of tasks completed correctly end-to-end. Each configuration was executed over multiple independent runs with a clean environment.

\begin{table}[ht]
\centering
\begin{tabular}{cccc}
\toprule
\textbf{Benchmark} & \textbf{Model} & \textbf{SR (W/O $\rightarrow$ W)} & \textbf{$\Delta$} \\
\midrule
\multirow{3}{*}{OAK} & GPT-OSS-120B    & $75\% \rightarrow 100\%$    & $+25$ pp \\
                     & GPT-4.1         & $70\% \rightarrow 96\%$     & $+26$ pp \\
                     & Claude Opus-4.5 & $81\% \rightarrow 96\%$     & $+15$ pp \\
\midrule
\multirow{3}{*}{BPO} & GPT-OSS-120B    & $49.2\% \rightarrow 82.3\%$ & $+33.1$ pp \\
                     & GPT-4.1         & $28.5\% \rightarrow 66.2\%$ & $+37.7$ pp \\
                     & Claude Opus-4.5 & $50.0\% \rightarrow 68.5\%$ & $+18.5$ pp \\
\bottomrule
\end{tabular}
\caption{Success Rate on OAK and BPO benchmarks, with and without CUGA's policy system, across three backbone models. W/O = without policy system; W = with policy system.}
\label{tab:all_results}
\end{table}


The results in Table~\ref{tab:all_results} show consistent and substantial gains across both benchmarks and all three models, ranging from $+15$ to $+37.7$ percentage points. On OAK Bench, the policy system drives GPT-OSS-120B to a perfect 100\% SR, with GPT-4.1 and Claude Opus-4.5 reaching 96\%. This is particularly evident in complex, multi-step tasks where, without the policy system, the agent deviates from required procedures or invokes tools out of sequence, mirroring how the demo's workflow becomes a structured, reliable process only once the Playbook is active. On the more demanding BPO benchmark, absolute gains are even larger: GPT-4.1 improves by $+37.7$ pp despite its lower baseline, suggesting the policy system is especially impactful where the model's native instruction-following is weakest. Taken together, these results confirm that \emph{governance-by-construction} is both effective and model-agnostic, achieving strong and consistent performance without reliance on frontier LLMs. Runtime governance does, however, introduce token overhead, an inherent consequence of replacing probabilistic compliance with explicit, auditable instructions (full figures in Section~\ref{app:bpo_benchmark}). In enterprise deployments where a single policy violation may constitute a compliance breach or trigger a destructive operation, this tradeoff is well justified.

\subsection{BPO Policy Ablation}
\label{subsec:bpo_ablation}

To isolate the contribution of individual governance primitives, we ran CUGA on BPO under a controlled ablation: \emph{no policies}, \emph{2 policies}, and \emph{5 policies}, using \texttt{GPT-OSS-120B}. Each configuration was executed over three independent runs with a clean vector store (Milvus) to eliminate carryover effects. Policies were matched at runtime via lightweight triggers and injected into the agent loop without model fine-tuning. Success rate improves monotonically: 46.2\% (12.0/26) with no policies $\rightarrow$ 71.8\% with 2 policies $\rightarrow$ 78.2\% with 5 policies (+32.0 pp overall). Each step corresponds directly to the governance primitives introduced in Section~\ref{sec:method_overview}:

\textbf{(1) Capability Boundaries (Intent Guard).}Without policies, the agent attempted to answer unsupported queries by requesting irrelevant identifiers or invoking unrelated APIs. An explicit capability-boundary policy caused it to correctly decline out-of-scope questions, converting multiple tasks from 0/3 to 3/3 (e.g., Tasks 16, 19, 21--23).

\textbf{(2) Tool Guide.} Several failures stemmed from calling known-unreliable endpoints (returning 503 errors or malformed schemas). Augmenting tool descriptions with warning guidance caused the agent to re-route to stable APIs, stabilizing previously flaky tasks (e.g., Tasks 12, 14, 18).

\textbf{(3) Playbook (Structured Reasoning Constraint).} For multi-metric questions, the agent previously shortcut to summary endpoints and returned incorrect counts. A multi-API reasoning playbook enforced granular tool selection, resolving systematic count errors (e.g., Task 9) and restoring full correctness. Full per-task results and evaluation sub-metrics are provided in Section~\ref{app:bpo_benchmark} of the supplementary material.

\section{Conclusion - Our Vision}
As LLM-based agents become increasingly prevalent across industry, their autonomous operation in diverse and sensitive environments introduces critical challenges around safety, compliance, and predictability. Recent incidents involving data and privacy leakage in computer-using agents underscore the urgency of addressing these risks, particularly as deployment moves beyond controlled research settings into production environments handling real users and sensitive data. Our vision is grounded in the belief that capable agents and trustworthy agents should be one and the same. As generalist computer-using agents progress toward production readiness, a principled governance layer becomes not a limitation, but a prerequisite for responsible deployment. This work demonstrates one realization of that vision through CUGA's policy system, which provides structured mechanisms to constrain unwanted and unpredictable agent behavior without sacrificing generality or performance, but the other way around. CUGA as an open-source, enterprise-oriented agent supporting on-premises deployment, reflects the operational realities of organizations that cannot rely on third-party cloud infrastructure for sensitive workloads. Its strong benchmark performance across diverse agent evaluation settings (e.g., AppWorld \cite{trivedi2024appworld}, WebArena \cite{zhou2023webarena}, BPO-TA \cite{shlomov2025benchmarks}) and by our results in Section~\ref{sec:results}, further validates it as a credible and production-relevant baseline. These properties, combined with lessons drawn from prior work and companies' use cases, informed both our design choices and our understanding of where policy enforcement is most needed. Ultimately, our policy system is a means to a broader end: the development of generalist agents that organizations can deploy with confidence.

\bibliographystyle{ACM-Reference-Format}
\bibliography{bib}

\appendix

\section{Ablation Study: BPO Benchmark}
\label{app:bpo_benchmark}

This appendix provides the complete experimental results for the CUGA policy
evaluation on the BPO benchmark, including per-run statistics, all metric
breakdowns, per-task pass rates, and descriptions of each implemented policy.

\subsection{Aggregate Results Across Policy Configurations}
\label{app:bpo_aggregate}

Each configuration was evaluated over three independent runs on a clean Milvus
database. Table~\ref{tab:bpo_runs} reports per-run scores and summary statistics.

\begin{table}[ht]
\centering
\begin{tabular}{lccc}
\toprule
 & \textbf{No Policies} & \textbf{2 Policies} & \textbf{5 Policies} \\
\midrule
Run 1 & 11/26 (42.3\%) & 19/26 (73.1\%) & 20/26 (76.9\%) \\
Run 2 & 13/26 (50.0\%) & 18/26 (69.2\%) & 20/26 (76.9\%) \\
Run 3 & 12/26 (46.2\%) & 19/26 (73.1\%) & 21/26 (80.8\%) \\
\midrule
Mean  & 12.0/26 (46.2\%) & 18.7/26 (71.8\%) & 20.3/26 (78.2\%) \\
Std.\ Dev. & 1.0 & 0.6 & 0.6 \\
Range & 11--13 & 18--19 & 20--21 \\
\bottomrule
\end{tabular}
\caption{Per-run scores for each policy configuration (out of 26 tasks).}
\label{tab:bpo_runs}
\end{table}

Table~\ref{tab:bpo_metrics} summarizes all evaluation metrics across
configurations, with the total improvement from no-policy to five-policy
baseline.

\begin{table*}[ht]
\centering
\begin{tabular}{lcccc}
\toprule
\textbf{Metric} & \textbf{No Policies} & \textbf{2 Policies} & \textbf{5 Policies} & \textbf{$\Delta$ Total} \\
\midrule
Final score        & 12.0/26 (46.2\%) & 18.7/26 (71.8\%) & 20.3/26 (78.2\%) & +32.0 pp \\
Avg.\ similarity   & 0.539 & 0.602 & 0.643 & +0.104 \\
Keyword match      & 57.5\% & 73.1\% & 76.0\% & +18.5 pp \\
LLM judge (avg)    & 0.517 & 0.742 & 0.824 & +0.307 \\
LLM judge (binary) & 52.6\% & 78.2\% & 88.5\% & +35.9 pp \\
API accuracy       & 50.0\% & 66.7\% & 73.1\% & +23.1 pp \\
\bottomrule
\end{tabular}
\caption{Evaluation metrics by policy configuration.}
\label{tab:bpo_metrics}
\end{table*}

\subsection{Per-Task Breakdown}
\label{app:bpo_per_task}

Table~\ref{tab:bpo_tasks} reports pass counts (out of 3 runs) for each task
across the three configurations, together with the net improvement and task
category.

\begin{table}[ht]
\centering
\small
\begin{tabular}{clcccc}
\toprule
\textbf{Task} & \textbf{Category} & \textbf{No Pol.} & \textbf{2 Pol.} & \textbf{5 Pol.} & \textbf{$\Delta$} \\
\midrule
1  & Stable pass                       & 3/3 & 3/3 & 3/3 & +0 \\
2  & Stable pass                       & 3/3 & 3/3 & 3/3 & +0 \\
3  & Stable fail                       & 0/3 & 0/3 & 0/3 & +0 \\
4  & Stable pass                       & 3/3 & 3/3 & 3/3 & +0 \\
5  & Stable pass                       & 3/3 & 3/3 & 3/3 & +0 \\
6  & Stable fail                       & 0/3 & 0/3 & 0/3 & +0 \\
7  & Stable fail                       & 0/3 & 0/3 & 0/3 & +0 \\
8  & Stable pass                       & 3/3 & 3/3 & 3/3 & +0 \\
9  & Improved (Policy \#3)             & 0/3 & 0/3 & 3/3 & +3 \\
10 & Stable pass                       & 3/3 & 3/3 & 3/3 & +0 \\
11 & Stable pass                       & 3/3 & 3/3 & 3/3 & +0 \\
12 & Flaky                             & 2/3 & 3/3 & 2/3 & +0 \\
13 & Stable pass                       & 3/3 & 3/3 & 3/3 & +0 \\
14 & Improved (Policy \#2)             & 2/3 & 3/3 & 3/3 & +1 \\
15 & Stable fail                       & 0/3 & 0/3 & 0/3 & +0 \\
16 & Improved (Policy \#1)             & 0/3 & 3/3 & 3/3 & +3 \\
17 & Stable pass                       & 3/3 & 3/3 & 3/3 & +0 \\
18 & Improved (Policy \#2)             & 2/3 & 3/3 & 3/3 & +1 \\
19 & Improved (Policy \#1)             & 0/3 & 3/3 & 3/3 & +3 \\
20 & Stable pass                       & 3/3 & 3/3 & 3/3 & +0 \\
21 & Improved (Policy \#1)             & 0/3 & 3/3 & 3/3 & +3 \\
22 & Improved (Policy \#1)             & 0/3 & 3/3 & 3/3 & +3 \\
23 & Improved (Policy \#1)             & 0/3 & 3/3 & 3/3 & +3 \\
24 & Improved (Policy \#1/\#2, partial)& 0/3 & 1/3 & 2/3 & +2 \\
25 & Stable fail                       & 0/3 & 1/3 & 0/3 & +0 \\
26 & Improved (Policy \#4)             & 0/3 & 0/3 & 3/3 & +3 \\
\bottomrule
\end{tabular}
\caption{Per-task pass rates across policy configurations.}
\label{tab:bpo_tasks}
\end{table}

\subsection{Policy Descriptions}
\label{app:bpo_policies}

The five implemented policies are described below in order of application.

\paragraph{\textbf{Policy \#1 — API Capability Boundaries}}
\textit{Type: Playbook.
Trigger: keywords + natural language (threshold 0.65, priority 90).}

This policy teaches the agent to recognise when no available API can answer a
query. It enumerates supported and unsupported API capabilities and instructs
the agent to decline out-of-scope requests directly, rather than soliciting a
requisition ID or invoking irrelevant tools.

\textbf{Failure pattern addressed.} Without this policy the agent would request
a requisition ID, or call arbitrary APIs, for queries relating to job
descriptions, time-to-fill, geography filtering, SLA deadlines, funnel timing,
and job-card details — none of which are supported by any API.

\textbf{Tasks fixed.} 16, 19, 21, 22, 23 (all from 0/3 to 3/3).

\paragraph{\textbf{Policy \#2 — Error-Prone Tool Warnings}}
\textit{Type: Tool Guide.
Target: 19 error-prone tools (warning prepended to each description).}

This policy prepends a warning to the descriptions of the 19 known-unreliable
tools (those returning 503 errors, schema violations, or type mismatches). It
steers the agent towards the 13 reliable core tools and teaches graceful
recovery when an error tool is nonetheless invoked.

\textbf{Failure pattern addressed.} The agent would call tools such as
\texttt{funnel\_status} (503 error), \texttt{model\_registry} (incorrect data),
or \texttt{source\_recommendation\_summary} (incomplete shortcut) instead of
the correct granular APIs.

\textbf{Tasks fixed.} 12, 14, 18 (stabilised from 2/3 to 3/3); 24 (0/3 to
1--2/3, partial improvement).

\paragraph{\textbf{Policy \#3 — Multi-API Reasoning}}
\textit{Type: Playbook.
Trigger: keywords + natural language (threshold 0.65, priority 80).}

This policy instructs the agent to call multiple specific APIs for multi-metric
questions rather than relying on a single summary endpoint. It provides an
explicit mapping from question type to the correct tool and clarifies the
distinction between ``total requisitions used for computation''
(\texttt{definitions-and-methodology}) and ``similar requisitions analysed''
(\texttt{metadata-and-timeframe}).

\textbf{Failure pattern addressed.} The agent would use the summary shortcut
tool for multi-metric queries, or confuse which endpoint returns the requisition
count.

\textbf{Tasks fixed.} 9 (0/3 to 3/3; agent now correctly returns 1047 from
\texttt{definitions-and-methodology} rather than 40 from metadata-and-timeframe).

\paragraph{\textbf{Policy \#4 — Average vs.\ Total Calculations}}
\textit{Type: Playbook.
Trigger: keywords + natural language (threshold 0.65, priority 70).}

This policy teaches the agent that when a user asks for ``average'' or
``typical'' values it must compute a per-requisition average by dividing the
aggregate total by the count of similar requisitions, rather than returning the
raw total.

\textbf{Failure pattern addressed.} The agent would return the total candidate
count (2{,}913) when asked ``how many candidates do we usually get?'' instead of
computing the per-requisition average ($2{,}913 \div 40 \approx 73$).

\textbf{Tasks fixed.} 26 (0/3 to 3/3).

\paragraph{\textbf{Policy \#5 — Missing Requisition ID vs.\ Unsupported Query}}
\textit{Type: Playbook.
Trigger: keywords + natural language (threshold 0.60, priority 85).}

This policy helps the agent distinguish between ``I need a requisition ID to
answer this'' (answerable but missing context) and ``this cannot be answered
regardless of requisition ID'' (unsupported by any API). It reinforces
Policy~\#1 for edge cases where the triggering language differs.

\textbf{Failure pattern addressed.} The agent would request a requisition ID
even for queries that no API supports under any circumstances.

\textbf{Tasks fixed.} Overlaps with Policy \#1; provides reinforcement for
edge cases not caught by Policy \#1's triggers.

\subsection{Remaining Failing Tasks}
\label{app:bpo_failures}

Table~\ref{tab:bpo_failures} documents the six tasks that remain unsolved (or
partially unsolved) after all five policies are applied, together with a
diagnosis of each failure mode.

\begin{table*}[ht]
\centering
\begin{tabular}{clp{7.5cm}}
\toprule
\textbf{Task} & \textbf{Pass rate} & \textbf{Issue} \\
\midrule
3  & 0/3 & Agent provides incomplete source data (only LinkedIn; misses Dice and GitHub details). Requires richer answer synthesis. \\
\midrule
6  & 0/3 & Garbled percentages from the summary tool. May require further tool-description refinement. \\
\midrule
7  & 0/3 & LLM judge scores correct agent behaviour as 0. Failure is in the judge, not the agent. \\
\midrule
12 & 2/3 & Agent occasionally omits one data-source name. Attributed to LLM non-determinism. \\
\midrule
15 & 0/3 & Agent calls too few APIs and misidentifies negative-SLA skills. Complex multi-part question. \\
\midrule
24 & 2/3 & Agent sometimes invokes an error tool for an invalid requisition ID. Partially addressed by Policy \#2. \\
\midrule
25 & 0/3 & Agent calls \texttt{requisition-details} error tool for requisition \texttt{UZLXBR} instead of declining. Edge case not caught by Policy \#1 triggers. \\
\bottomrule
\end{tabular}
\caption{Remaining failing tasks after five-policy configuration.}
\label{tab:bpo_failures}
\end{table*}

\section{Policy System Evaluations}
We presented CUGA's policy system evaluation in Section~\ref{sec:results}. Here, we extend that discussion by reporting the token usage associated with each evaluation, providing a quantitative measure of the overhead introduced by our method. Results are shown in Table~\ref{tab:results_with_cost}.

\begin{table*}[ht]
\centering
\begin{tabular}{ccccc}
\toprule
\textbf{Benchmark} & \textbf{Model} & \textbf{Accuracy (W/O $\rightarrow$ W)} & \textbf{$\Delta$} & \textbf{Tokens (W/O $\rightarrow$ W)} \\
\midrule
\multirow{3}{*}{OAK} & GPT-OSS-120B    & $75\% \rightarrow 100\%$ & $+25$ pp & $26.2k \rightarrow 40k$                                 \\
                     & GPT-4.1         & $70\% \rightarrow 96\%$  & $+26$ pp & $28.2k \rightarrow 44k$  \\
                     & Claude Opus-4.5 & $81\% \rightarrow 96\%$  & $+15$ pp & $44k \rightarrow 85k$    \\
\midrule
\multirow{4}{*}{BPO} & GPT-OSS-120B    & $49.2\% \rightarrow 82.3\%$ & $+33.1$ pp & $3.7M \rightarrow 6.6M$  \\
                     & GPT-4.1         & $28.5\% \rightarrow 66.2\%$ & $+37.7$ pp & $2.7M \rightarrow 8.8M$  \\
                     & Claude Opus-4.5 & $50.0\% \rightarrow 68.5\%$ & $+18.5$ pp & $6.2M \rightarrow 9.0M$ \\
\bottomrule
\end{tabular}
\caption{Success Rate and Token Usage on OAK and BPO benchmarks, with and without CUGA's policy system, across three backbone models. W/O = without policy system; W = with policy system.}
\label{tab:results_with_cost}
\end{table*}

\end{document}